\documentclass{article}
\usepackage{makecell}
\usepackage{graphicx}
\usepackage{booktabs}
\usepackage{multirow}
\usepackage{float}
\usepackage{subcaption}
\usepackage{xcolor}
\usepackage[table]{xcolor}
\usepackage{wrapfig}
\newcommand{\hlcell}[1]{\cellcolor{gray!15}#1}
\usepackage[preprint]{neurips_2026}

\usepackage[utf8]{inputenc} % allow utf-8 input
\usepackage[T1]{fontenc}    % use 8-bit T1 fonts
\usepackage{hyperref}       % hyperlinks
\usepackage{url}            % simple URL typesetting
\usepackage{booktabs}       % professional-quality tables
\usepackage{amsfonts}       % blackboard math symbols
\usepackage{nicefrac}       % compact symbols for 1/2, etc.
\usepackage{microtype}      % microtypography
\usepackage{xcolor}         % colors
\usepackage{amsmath}
\usepackage{algorithm}
\usepackage{algpseudocode}
\title{Trajectory-Consistent Calibration for Cache-Accelerated Diffusion Models}

\author{
  Mingyu Liang\thanks{Equal contribution.}\quad
  Dingkun Xu\footnotemark[1]\quad
  Jingwei Xu\thanks{Corresponding author.}\\
  Laboratory for Novel Software Technology, Nanjing University, China\\
  \texttt{\{522025330055, 502025330057\}@smail.nju.edu.cn}\\
  \texttt{jingweix@nju.edu.cn}
}

\begin{document}

\maketitle

\begin{abstract}
Diffusion Transformers require repeated denoiser evaluations during iterative sampling, making inference computationally expensive.
Cache-based acceleration reduces this cost by reusing intermediate representations across denoising steps, but can introduce representation deviations and degrade generation quality.
In this paper, we analyze these deviations and show that effective calibration should consider both the direct mismatch caused by reuse and the subsequent trajectory shift induced by earlier corrections.
To address this challenge, we propose \emph{Trajectory-Consistent Calibration} (TCC), a training-free method that calibrates cached representations toward their full-computation counterparts.
Specifically, rather than estimating all calibration priors from a single uncorrected cache trajectory, TCC uses an offline iterative procedure so that each prior accounts for the trajectory shift induced by preceding calibrations.
Experiments on PixArt-$\alpha$ and DiT-XL/2 show that TCC consistently improves FID across representative cache-based acceleration methods while preserving their underlying reuse policies.
Notably, in a representative PixArt-$\alpha$ cache-acceleration setting based on FORA, TCC reduces FID from 29.83 to 27.35, slightly surpassing the full-computation baseline.

\end{abstract}

\section{Introduction}
\begin{figure}[ht]
    \centering
    \includegraphics[width=\linewidth]{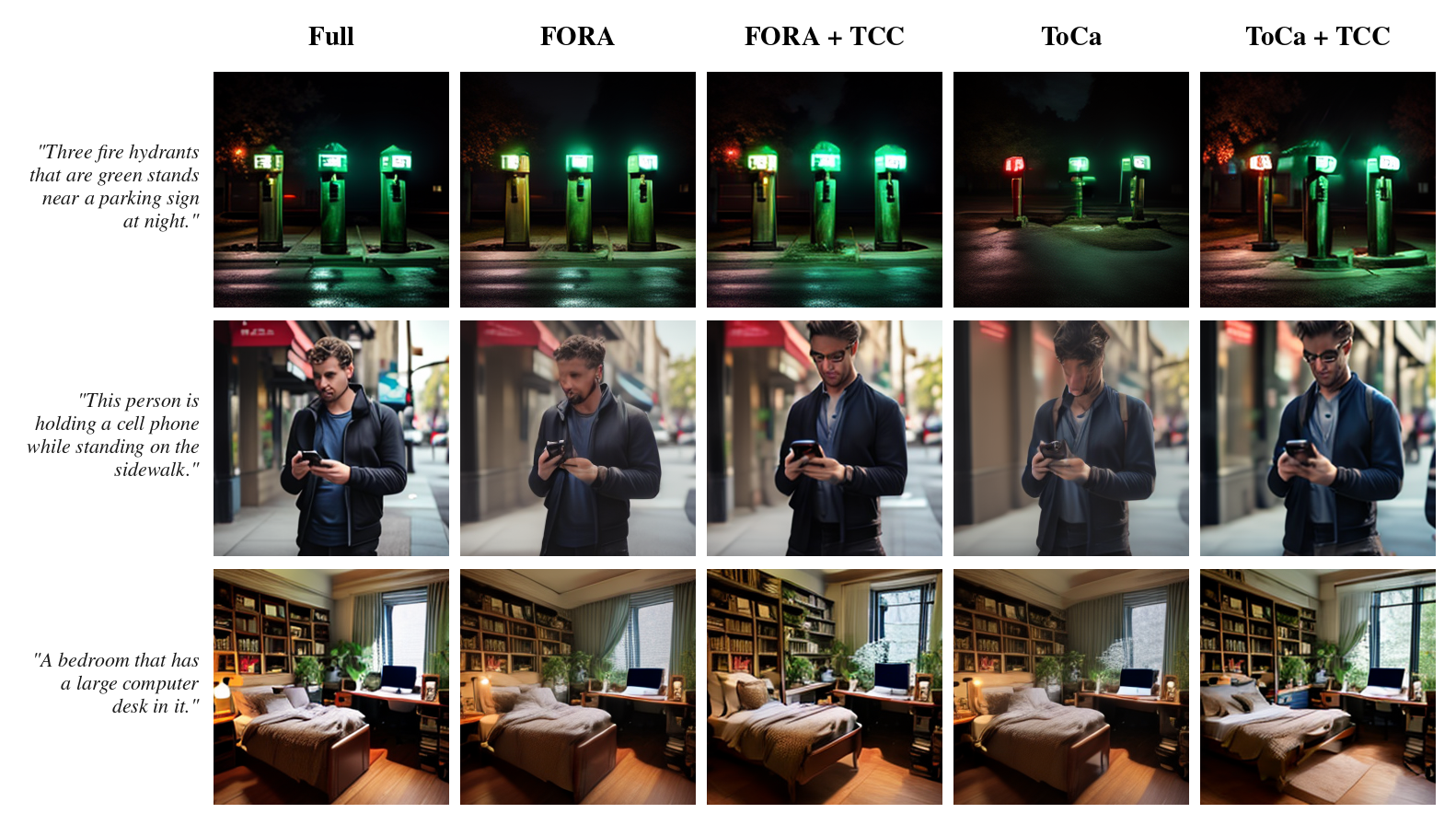}
\caption{
Qualitative comparison on PixArt-$\alpha$ under cache-accelerated sampling, with prompts shown verbatim.
TCC recovers prompt-relevant content weakened by cache reuse, such as the parking sign and green fire hydrants, and mitigates detail degradation when semantics are preserved, e.g., facial details.
The third row shows that TCC preserves visual quality under complex compositions.
All comparisons use matched prompts, seeds, and sampling settings.
}

    \label{fig:pixart_qualitative}
\end{figure}

Diffusion Transformers (DiTs)~\citep{dit} have substantially advanced visual generation by bringing scalable transformer backbones to diffusion models. However, their iterative denoising procedure requires repeated forward passes of a large backbone, making inference computationally expensive. Since generating a single sample involves multiple denoising steps~\citep{ddpm, ddim}, reducing repeated computation has become critical for efficient diffusion sampling. 

Cache-based acceleration methods address this challenge by caching and reusing intermediate representations across nearby denoising steps in DiT.
By exploiting temporal redundancy along the diffusion trajectory, they reduce repeated computation and provide a practical training-free route to faster inference~\citep{deepcache,blockcache}.
However, cache reuse is inherently approximate rather than lossless. Because a reused representation is computed from a previous or neighboring denoising state~\citep{deepcache}, it can deviate from the representation that full computation would produce at the current state. As sampling proceeds, this approximation error can further propagate through later denoising steps, causing the cache-accelerated trajectory to drift from the full-computation trajectory and ultimately degrade generation quality.

A key observation of this work is that the deviation introduced by cache reuse is not purely local. Instead, effective calibration for cache reuse should account for two coupled effects: the \emph{direct mismatch} between a reused representation and its full-computation counterpart at the current denoising step, and the \emph{trajectory shift} induced when earlier reuse and calibration alter the subsequent cache-side trajectory. Existing cache acceleration methods mainly improve reuse itself. 
One line of work focuses on reuse decisions, designing or learning cache schedules that determine which timesteps, layers, tokens, or states can be safely reused~\citep{fora, l2c, toca}. 
Another line introduces optimization, calibration, or compensation mechanisms to better approximate the computation skipped by caching~\citep{2025icc,2025goc}.
However, these approaches do not explicitly model the coupled effect of direct mismatch and the trajectory shift introduced by reuse correction.
Prior correction methods typically estimate a fixed error prior from the original cache and full-computation trajectories.
In contrast, TCC treats calibration as part of the trajectory and estimates each subsequent prior under the corrected history induced by previous calibrations.

To address this challenge, we propose \emph{Trajectory-Consistent Calibration} (TCC), a training-free method that calibrates cache-induced direct mismatch and trajectory shift without modifying the underlying cache schedule or model parameters.
TCC takes the full-computation trajectory as a reference and estimates calibration priors from paired cache-side and full-computation representations to correct direct mismatch at each calibration stage.
To account for trajectory shift induced by earlier reuse and calibration, these priors are estimated offline and iteratively along the corrected cache trajectory, so that each stage matches the history shaped by preceding corrections.
This trajectory-consistent prior estimation preserves the base reuse policy with only a small transformation cost.

We validate TCC on PixArt-$\alpha$ and DiT-XL/2 across representative cache baselines, including FORA, ToCa, and L2C.
TCC consistently improves FID in both text-to-image and class-conditional generation, with representative gains from 29.83 to 27.35 on PixArt-$\alpha$ and from 6.82 to 5.70 on DiT-XL/2. Qualitative examples are shown in Figure~\ref{fig:pixart_qualitative} and Appendix~\ref{app:qualitative}.
Our contributions are summarized as follows:
\begin{itemize}
\item We formulate the cache-induced degradation in diffusion transformers as arising from two coupled effects: direct mismatch between cache-side and full-computation representations, and trajectory shift induced by preceding reuse and calibration.
    
\item We propose \emph{Trajectory-Consistent Calibration} (TCC), a training-free method that jointly handles these two effects via statistical calibration and iterative trajectory-consistent prior estimation without modifying the cache strategy or model parameters.

\item We validate TCC on PixArt-$\alpha$ and DiT-XL/2 with representative cache baselines. TCC consistently improves FID, including in aggressive-reuse settings, and ablations verify the stability advantage of trajectory-consistent prior estimation over one-shot calibration.
\end{itemize}

\section{Problem Formulation}
Diffusion sampling generates a sample by progressively updating a sequence of denoising states \(x_T, x_{T-1}, \ldots, x_0\). At each timestep \(t\), a transformer-based denoiser performs a forward pass on the current state \(x_t\) to produce the prediction used for the next update. During this forward pass, the denoiser produces intermediate representations at multiple internal sites. We denote such a site $u$ by 
\[
u=(t,\ell,m),
\]
where \(t\) is the timestep, \(\ell\) is a transformer layer, and \(m\) specifies a module output site, such as self-attention, cross-attention, or MLP outputs in transformer-based denoisers. 

Let \(h_u\) denote the output representation at site \(u\), with \(h_{<u}\) denoting the local incoming representation immediately before this site. Let \(\mathcal{F}_u\) denote the fresh computation performed at site \(u\), which maps \(h_{<u}\) to \(h_u\). We use the superscript \(\mathrm{full}\) to denote quantities obtained along the full-computation trajectory, where all denoiser computations are freshly evaluated. The full-computation representation at site \(u\) is then written as
\[
h_u^{\mathrm{full}}=\mathcal{F}_u(h_{<u}^{\mathrm{full}}).
\]

We model an existing cache-based accelerator as a cache strategy \(\pi\) that specifies how each intermediate site is evaluated. Under this formulation, \(\pi\) is represented by a set of cache-affected sites \(\mathcal{S}_\pi\) and site-wise reuse or approximation operators \(\{\Phi_u^\pi\}_{u\in\mathcal{S}_\pi}\), both of which may depend on the preceding trajectory history for adaptive cache strategies. To describe this history dependence, let \(\mathcal{H}_{<u}^{\mathrm{full}}\) denote the trajectory history reached before site \(u\) when all preceding computations are freshly evaluated, and let \(\mathcal{H}_{<u}^{\pi}\) denote the trajectory history reached before site \(u\) when the base cache strategy \(\pi\) is applied along the preceding trajectory without additional calibration. At site \(u\), the cache-side representation under \(\pi\), with local incoming representation \(h_{<u}^{\pi}\), is written as
\[
h_u^\pi =
\begin{cases}
\mathcal{F}_u(h_{<u}^\pi), & u\notin \mathcal{S}_\pi,\\
\Phi_u^\pi(h_{<u}^\pi,\mathcal{H}_{<u}^{\pi}), & u\in \mathcal{S}_\pi.
\end{cases}
\]
The realization of \(\mathcal{S}_\pi\) and \(\Phi_u^\pi\) is determined by the base cache strategy.
Given the base cache strategy \(\pi\), we select a set of calibration sites \(\mathcal{S}_{\mathrm{cal}}\subseteq\mathcal{S}_\pi\) and construct calibration operators \(\mathcal{C}=\{\mathcal{C}_u\}_{u\in\mathcal{S}_{\mathrm{cal}}}\) for these selected cache-affected sites. Cache-affected sites outside \(\mathcal{S}_{\mathrm{cal}}\) follow the base cache strategy without an additional calibration operator.

\textbf{Direct representation mismatch.}
For a cache-affected site \(u\in\mathcal{S}_\pi\), the cache-side representation \(h_u^\pi\) is generally an approximation to its full-computation counterpart \(h_u^{\mathrm{full}}\), as \(\Phi_u^\pi\) replaces the fresh computation \(\mathcal{F}_u\). This discrepancy can also reflect preceding cache decisions, which may make the incoming representation \(h_{<u}^{\pi}\) differ from \(h_{<u}^{\mathrm{full}}\). We refer to the resulting discrepancy between \(h_u^\pi\) and \(h_u^{\mathrm{full}}\) as direct representation mismatch. At selected calibration sites, \(\mathcal{C}_u\) is applied to reduce this direct mismatch before the representation is consumed by subsequent denoising computation.

\textbf{Trajectory-induced prior mismatch.}
The second mismatch concerns the trajectory on which calibration priors are estimated and applied. A calibration operator estimated from cache-side representations collected under one trajectory history may no longer match its application context after earlier calibrations have changed the subsequent cache-side trajectory. We use \(\mathcal{H}_{<u}^{\mathrm{corr}}\) to denote the history reached before site \(u\) after previous calibration operators have been applied along the same base cache strategy \(\pi\). For \(u\in\mathcal{S}_{\mathrm{cal}}\), with local incoming representation \(h_{<u}^{\mathrm{corr}}\), the representation passed to the subsequent denoising computation is
\[
h_u^{\mathrm{corr}}
=
\mathcal{C}_u\!\left(
\Phi_u^\pi(h_{<u}^{\mathrm{corr}},\mathcal{H}_{<u}^{\mathrm{corr}})
\right).
\]
Because calibrated representations propagate forward, later calibration operators are applied under trajectory histories shaped by preceding calibrations. If the corresponding priors are instead estimated without accounting for these preceding calibrations, they can become mismatched to their application contexts. This motivates trajectory-consistent prior estimation, where each prior is collected along the corrected trajectory on which it will be applied.
Accordingly, TCC addresses these two effects with local statistical calibration and trajectory-consistent prior estimation, respectively.

\section{Method}
\begin{figure}[t]
  \centering
  \includegraphics[width=\linewidth]{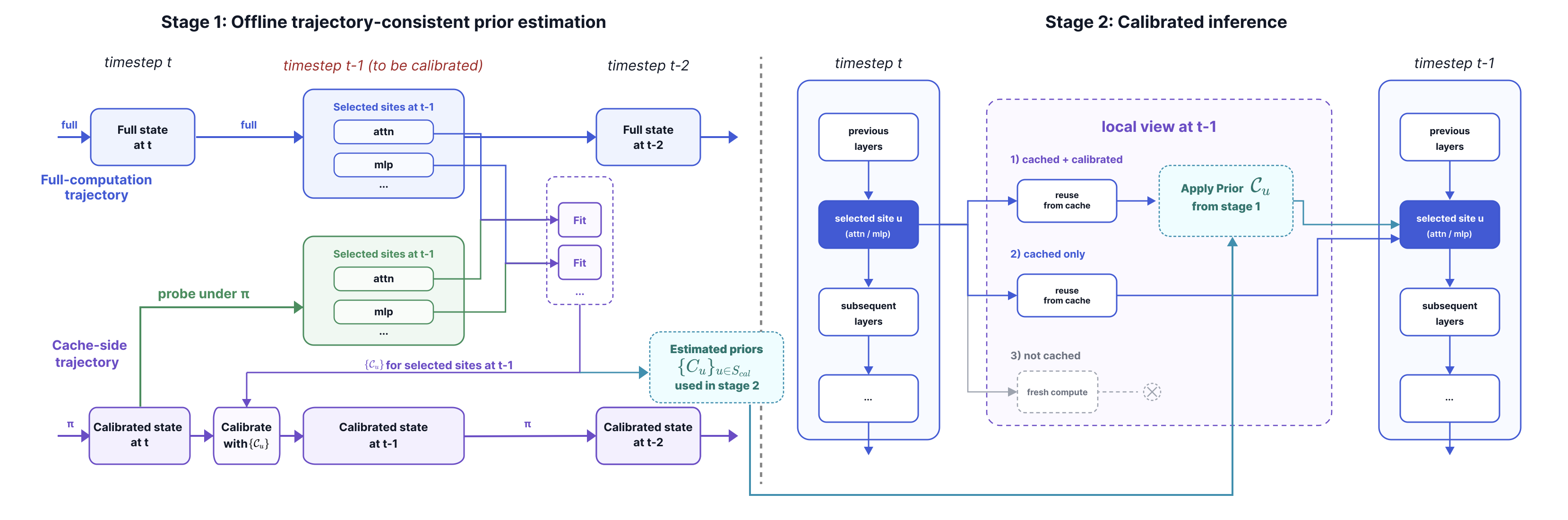}
\caption{
Overview of trajectory-consistent calibration (TCC).
In Stage~1, at each calibration timestep, TCC probes selected calibration sites, e.g., attention and MLP outputs, under cache strategy $\pi$, estimates calibration transforms from cache-side to full-computation representations, and performs a calibrated advance so that subsequent priors are collected on the trajectory induced by previous calibrations.
In Stage~2, each selected reused representation is calibrated before propagation to subsequent layers, while non-calibrated sites follow the base cache strategy.
}
  \label{fig:method_overview}
\end{figure}

This section presents \emph{Trajectory-Consistent Calibration} (TCC), a training-free calibration method for cache-accelerated diffusion models. Given a cache strategy \(\pi\), TCC operates on selected calibration sites \(\mathcal{S}_{\mathrm{cal}}\subseteq\mathcal{S}_\pi\), without modifying the denoiser, sampler, or cache strategy. As illustrated in Figure~\ref{fig:method_overview}, TCC consists of two stages. In the offline prior estimation stage, it estimates local calibration priors from paired full-computation and cache-side representations along the corrected cache trajectory, so that each prior is matched to the trajectory history under which it will later be applied. In the calibrated inference stage, the cache strategy first produces a cache-side representation, and the corresponding local calibration operator is then applied before the representation is consumed by subsequent denoising computation. We next define the local calibration operator and the trajectory-consistent prior estimation procedure.

\subsection{Local Statistical Calibration}
For a cache-affected site \(u\in\mathcal{S}_{\mathrm{cal}}\), our goal is to construct a local calibration operator that maps the cache-side representation toward its full-computation counterpart. We model the direct representation mismatch as a statistical transformation from cache-side representations to their full-computation representations. Our key assumption is that, given a site \(u\), these two sets of representations exhibit a systematic statistical shift that can be estimated from representative samples.

Given \(N\) representative samples indexed by \(i\), let \(a_i=h_{u,i}^{\mathrm{full}}\) denote the full-computation output representation at site \(u\), and let \(b_i\) denote the corresponding cache-side output representation, whose collection is described in Section~3.2. We collect the paired representations
\[
\{(a_i,b_i)\}_{i=1}^{N}.
\]
We stack these paired vectors as rows of
\[
A=[a_1;\ldots;a_N],\qquad
B=[b_1;\ldots;b_N],
\]
which form the full-computation and cache-side representation matrices, with each row indexed by $i$ providing one paired representation vector. In our implementation, for class-conditioned DiT, we average representation vectors from several sampled trajectories of the same class; for text-conditioned PixArt-\(\alpha\), each text prompt serves as one representative sample.

We instantiate this local operator with a closed-form rotation-scale fit based on paired representation statistics, inspired by scaled Orthogonal Procrustes alignment~\citep{schonemann1966procrustes}. This fit uses first-order moments to correct mean shifts and second-order cross-covariance statistics to estimate the geometric mismatch between the cache-side and full-computation representation distributions. The resulting local calibration operator is closed-form and training-free: it centers cache-side representations by their own mean, rotates them toward the full-computation representation geometry, rescales their magnitude, and then shifts them to the full-computation mean.

Specifically, with \(\mu_A\) and \(\mu_B\) denoting the row means of \(A\) and \(B\), we center the representation as
\[
A_c=A-\mu_A,\qquad
B_c=B-\mu_B.
\]
We then estimate an orthogonal rotation \(R\) and a non-negative scale \(s\) by solving a scaled Orthogonal Procrustes problem:
\[
(R,s)
=
\underset{R^\top R=I,\;s\ge 0}{\operatorname{arg\,min}}\;
\left\|A_c-sB_cR\right\|_F^2.
\]

We compute the singular value decomposition of the cross-covariance matrix as $B_c^\top A_c = U\Sigma V^\top.$ The Procrustes rotation and scale are then obtained as
\[
\begin{aligned}
R &= UV^\top,\\
s &=
\frac{\langle A_c,B_cR\rangle}
{\|B_cR\|_F^2+\epsilon}.
\end{aligned}
\]
The corresponding full statistical transform is
\[
\mathcal{T}(h)=\mu_A+s(h-\mu_B)R,
\]
where \(R\) rotates the centered cache-side representation and \(s\) rescales it before shifting it to the full-computation mean. Combining mean alignment, rotation, and scale, we define the fitting rule \(\operatorname{Fit}(A,B;\alpha)\) that returns the local calibration operator \(\mathcal{C}\):
\begin{equation}
\mathcal{C}=\operatorname{Fit}(A,B;\alpha),
\qquad
\mathcal{C}(h)
=
h+\alpha\bigl(\mathcal{T}(h)-h\bigr).
\label{eq:fitting-rule}
\end{equation}
Here \(\alpha\) controls the calibration strength, and the returned function \(\mathcal{C}\) is applied to cache-side representations during calibrated inference.
When \(\alpha=0\), \(\mathcal{C}\) reduces to the identity map, so TCC reduces to the base cache strategy. When \(\alpha=1\), \(\mathcal{C}\) applies the full statistical calibration transform. Intermediate values interpolate between the uncalibrated cache-side representation and its calibrated representation. We treat \(\alpha\) as an experimental hyperparameter and analyze its effect empirically.

In practice, for token or spatial dimensions, we obtain representative vectors through lightweight pooling to keep prior estimation compact. The fitting rule above returns the calibration operator used at the fixed calibration site \(u\); in the site-indexed notation of the formulation, this operator is written as \(\mathcal{C}_u\). The next subsection describes how these paired batches are formed under corrected trajectory histories.

\subsection{Trajectory-Consistent Prior Estimation}

Given a timestep \(t\), we fit local calibration operators for the selected sites \(u\) in this timestep using the fitting rule defined in Section~3.1. The prior-estimation procedure advances the full-computation and corrected cache trajectories timestep by timestep. When it reaches timestep \(t\), the corrected cache trajectory already contains the effect of earlier calibrated timesteps. Therefore, the calibration priors for selected sites in timestep \(t\) should be estimated from cache-side representations produced under the trajectory history in which the corresponding calibration operators will be applied.

For representative samples \(i=1,\ldots,N\), and for each selected site \(u=(t,\ell,m)\in\mathcal{S}_{\mathrm{cal}}\) in timestep \(t\), we instantiate the paired matrices \(A\) and \(B\) in Section~3.1 as the site-specific batches
\[
\begin{aligned}
A_u^{\mathrm{full}}
&=
\left[
\mathcal{F}_{u}
\left(
h_{<u,i}^{\mathrm{full}}
\right)
\right]_{i=1}^{N}
=
[h_{u,i}^{\mathrm{full}}]_{i=1}^{N},
\\
B_u^{\mathrm{probe}}
&=
\left[
\Phi_u^\pi
\left(
h_{<u,i}^{\mathrm{corr}},
\mathcal{H}_{<u,i}^{\mathrm{corr}}
\right)
\right]_{i=1}^{N}.
\end{aligned}
\]
Here \(A_u^{\mathrm{full}}\) is collected from the full-computation trajectory. The cache-side batch \(B_u^{\mathrm{probe}}\) is collected by advancing a probe cache-side trajectory through timestep \(t\) from the current corrected history using \(\pi\), while recording the selected sites in this timestep without updating \(\mathcal{H}^{\mathrm{corr}}\). The probe pass records the cache-side representations used to fit $C_u$ but does not commit its output to $H^{\mathrm{corr}}$; only the following calibrated advance updates $H^{\mathrm{corr}}$. A calibration prior estimated without accounting for preceding calibrations would generally be mismatched to the cache-side representations encountered during calibrated inference. TCC therefore estimates each calibration prior under its inference-time trajectory history.
Using the fitting rule defined in Equation~\ref{eq:fitting-rule}, the local calibration operator at site \(u\) is fitted from \(A_u^{\mathrm{full}}\) and \(B_u^{\mathrm{probe}}\):
\[
\mathcal{C}_{u}
=
\operatorname{Fit}
\left(
A_u^{\mathrm{full}},
B_u^{\mathrm{probe}};\alpha
\right).
\]
After the local operators for timestep \(t\) are fitted, TCC performs a calibrated cache-side advance through timestep \(t\) from the same incoming corrected history. During this calibrated advance, each \(\mathcal{C}_u\) is applied at its corresponding selected site before the representation is consumed by subsequent denoising computation, and the timestep update advances \(\mathcal{H}^{\mathrm{corr}}\). This trajectory-consistent prior estimation procedure is summarized in Algorithm~\ref{alg:seq_collection}.

\newcommand{\StateBlock}[1]{%
    \State \begin{tabular}[t]{@{}l@{}}#1\end{tabular}%
}

\begin{algorithm}[t]
\caption{Offline trajectory-consistent prior estimation}
\label{alg:seq_collection}
\begin{algorithmic}[1]
\Require Cache strategy \(\pi\), calibration sites \(\mathcal{S}_{\mathrm{cal}}\), representative samples, calibration strength \(\alpha\)
\State Initialize \(\mathcal{H}^{\mathrm{full}}\) and \(\mathcal{H}^{\mathrm{corr}}\) on the same samples
\For{each timestep \(t\)}
    \If{timestep \(t\) contains selected calibration sites}
        \StateBlock{Advance the full-computation trajectory through timestep \(t\) from \(\mathcal{H}^{\mathrm{full}}\),\\
        \quad recording \(A_u^{\mathrm{full}}\) for selected sites \(u\) in this timestep and updating \(\mathcal{H}^{\mathrm{full}}\)}
        \StateBlock{Advance a probe cache-side trajectory through timestep \(t\) from \(\mathcal{H}^{\mathrm{corr}}\) using \(\pi\),\\
        \quad recording \(B_u^{\mathrm{probe}}\) for selected sites \(u\) in this timestep \textbf{without updating} \(\mathcal{H}^{\mathrm{corr}}\)}
        \For{each selected site \(u\) in this timestep}
            \State Estimate \(\mathcal{C}_{u}\leftarrow\operatorname{Fit}(A_u^{\mathrm{full}},B_u^{\mathrm{probe}};\alpha)\)
        \EndFor
        \StateBlock{Advance the cache-side trajectory through timestep \(t\) from \(\mathcal{H}^{\mathrm{corr}}\) using \(\pi\),\\
        \quad applying \(\mathcal{C}_u\) at selected sites in this timestep and updating \(\mathcal{H}^{\mathrm{corr}}\)}
    \Else
        \State Advance and update \(\mathcal{H}^{\mathrm{full}}\) through timestep \(t\) under full computation
        \State Advance and update \(\mathcal{H}^{\mathrm{corr}}\) through timestep \(t\) using the cache strategy \(\pi\)
    \EndIf
\EndFor
\State \Return \(\{\mathcal{C}_{u}\}_{u\in\mathcal{S}_{\mathrm{cal}}}\)
\end{algorithmic}
\end{algorithm}

Consequently, later timesteps are reached under the corrected history updated by earlier calibrated timestep advances. TCC therefore preserves site-local calibration operators while estimating their priors under the corrected trajectory histories induced by preceding calibrations.

\section{Experiments}
\subsection{Experiment Settings}

\textbf{Model configurations.}
We evaluate TCC on two representative generation tasks: text-to-image generation
and class-conditional image generation, using NVIDIA H800 80GB GPUs. For text-to-image generation, we use
PixArt-$\alpha$~\citep{pixartalpha} with DPM-Solver++~\citep{dpmsolverplusplus} sampling at
$256\times256$ resolution and 20 sampling steps. We compare TCC with
representative training-free cache-based acceleration baselines, including FORA~\citep{fora}
and ToCa~\citep{toca}. We do not include L2C~\citep{l2c} in this setting, since the original L2C work
trains and evaluates its router on class-conditional DiT models.

For class-conditional generation, we adopt DiT-XL/2~\citep{dit} with the DDIM~\citep{ddim} sampler and
evaluate two 20-step settings: $256\times256$ and $512\times512$ generation. On
DiT-XL/2, we compare with representative cache-based acceleration methods,
including the learned-policy L2C and the training-free FORA. We further compare
with EOC~\citep{eoc}, a recent training-free cache-optimization method that also aims to
mitigate cache-induced quality degradation, but does so through a different
mechanism from TCC.

\textbf{Prior statistics and evaluation.}
TCC estimates calibration statistics offline and applies the resulting
calibration transforms during cached sampling. For PixArt-$\alpha$, we construct
a prompt-level prior set using 1K prompts sampled from the MS-COCO-2017~\citep{coco} training
captions, without overlap with the 30K prompts used for FID~\citep{fid} evaluation. For
DiT-XL/2, we construct class-wise priors on ImageNet~\citep{imagenet} by generating 100 samples
per class and averaging token-pooled representations to obtain class-level
representative statistics. All prior statistics and calibration matrices are computed in FP32.
Offline prior estimation is performed once for each model and base cache setting and is therefore not included in per-sample inference latency; the reported latency includes the calibration transforms applied during cached sampling.
We report both FLOPs and wall-clock latency because dynamic token selection and memory movement can make FLOPs an incomplete proxy for end-to-end speed.
We provide a diagnostic analysis of the TCC application window
in Appendix~\ref{app:within_label_dispersion}.

For text-to-image generation, we follow the MS-COCO-30K protocol used by ToCa
and report FID-30K for image quality and CLIP Score for image-text alignment.
The CLIP Score is computed using CLIP ViT-B/32~\citep{clip}. For class-conditional generation,
we follow the standard DiT evaluation protocol by uniformly sampling 50,000
images from the 1,000 ImageNet classes. We report FID-50K, sFID, and Inception
Score~\citep{is}. 
For each base cache strategy and evaluation setting, we use a fixed calibration window, site set, and correction strength for all generated samples, rather than adapting these choices per sample.
Further details on hyperparameter selection are provided in Appendix~\ref{app:hyperparameter_selection}.

\textbf{Code availability.}
Our code is available at \url{https://github.com/NJUDeepEngine/TCC}.

\subsection{Main Results}

\paragraph{Text-to-image generation.}
\begin{table}[t]
\centering
\small
\setlength{\tabcolsep}{5pt}
\caption{
Text-conditional image generation on MS-COCO 2017 with PixArt-$\alpha$.
Latency is measured end-to-end, including ToCa's dynamic token weighting and TCC calibration; hyperparameters are in Appendix~\ref{app:hyperparameter_selection}.
}
\label{tab:pixart_coco}
\begin{tabular}{lccc|cc}
\toprule
\multirow{2}{*}{\textbf{Method}}
& \multirow{2}{*}{\textbf{Latency (s)} $\downarrow$}
& \multirow{2}{*}{\textbf{FLOPs} $\downarrow$}
& \multirow{2}{*}{\textbf{Speed} $\uparrow$}
& \multicolumn{2}{c}{\textbf{MS-COCO2017}} \\
& & & &
\textbf{FID-30K} $\downarrow$ &
\textbf{CLIP} $\uparrow$ \\
\midrule
PixArt-$\alpha$
& 0.9832 & 11.1225 & 1.00$\times$ & 28.15 & 31.03 \\
50\% steps
& 0.4997 & 5.5613 & 1.97$\times$ & 37.47 & 30.50 \\
\midrule
FORA ($N=3$)
& 0.4580 & 3.8902 & 2.15$\times$ & 29.83 & 30.98 \\
\rowcolor{gray!15}
FORA ($N=3$) + TCC
& 0.5388 & 4.3469 & 1.82$\times$ & \textbf{27.35} & 30.75 \\
\midrule
ToCa ($N=3$, $R=60\%$)
& 0.9720 & 6.5357 & 1.01$\times$ & 28.03 & 30.95 \\
ToCa ($N=3$, $R=90\%$)
& 0.9210 & 4.6580 & 1.07$\times$ & 29.63 & 30.95 \\
\rowcolor{gray!15}
ToCa ($N=3$, $R=90\%$) + TCC
& 1.0480 & 5.1146 & 0.94$\times$ & \textbf{27.79} & 30.33 \\
\bottomrule
\end{tabular}
\end{table}

Table~\ref{tab:pixart_coco} reports the results on PixArt-$\alpha$. TCC
improves FID-30K for both FORA and ToCa. For FORA with $N=3$, TCC reduces
FID-30K from 29.83 to 27.35, improving over both the cache baseline and the
full-computation PixArt-$\alpha$ baseline. For ToCa, the aggressive $R=90\%$
setting lowers FLOPs compared with the milder $R=60\%$ setting, but increases
FID-30K from 28.03 to 29.63. Applying TCC to this aggressive setting reduces
FID-30K to 27.79, suggesting that TCC can recover cache-induced quality
degradation under stronger reuse.
This gain is mainly reflected in FID: CLIP Score remains close for FORA+TCC but decreases under the aggressive ToCa+TCC setting, indicating that calibration does not uniformly improve all evaluation metrics.
The lower wall-clock speedup of ToCa+TCC
reflects ToCa's dynamic token-level overhead and the added calibration cost;
a detailed breakdown is provided in Appendix~\ref{app:latency_overhead}. For
module-level cache methods such as FORA, TCC preserves clear wall-clock
acceleration while improving FID.

\textbf{Class-conditional image generation.}
\begin{table}[t]
\centering
\caption{
Results on ImageNet with DiT-XL/2 under 20-step DDIM sampling.
TCC hyperparameters are summarized in Appendix~\ref{app:hyperparameter_selection}.
}
\label{tab:imagenet_20_merged}
\small
\setlength{\tabcolsep}{4pt}
\begin{tabular}{llcccccc}
\toprule
Resolution & Method
& Latency (s) $\downarrow$
& FLOPs $\downarrow$
& Speed $\uparrow$
& FID $\downarrow$
& sFID $\downarrow$
& IS $\uparrow$ \\
\midrule
\multirow{7}{*}{$256\times256$}
& DiT
& 0.8164 & 9.5055 & 1.00$\times$ & 3.51 & 4.93 & 223.54 \\
\cmidrule(lr){2-8}
& L2C
& 0.6704 & 7.4006 & 1.22$\times$ & 3.50 & 4.66 & \textbf{225.44} \\
& L2C + EOC
& 0.7249 & 7.4100 & 1.12$\times$ & 3.45 & 4.68 & 223.96 \\
& \hlcell{L2C + TCC}
& \hlcell{0.7127}
& \hlcell{7.6887}
& \hlcell{1.15$\times$}
& \hlcell{\textbf{3.45}}
& \hlcell{\textbf{4.62}}
& \hlcell{224.54} \\
\cmidrule(lr){2-8}
& FORA ($N=2$)
& 0.4727 & 4.7634 & 1.73$\times$ & 6.82 & 8.65 & 190.01 \\
& FORA ($N=2$) + EOC
& 0.5533 & 4.7700 & 1.48$\times$ & 5.82 & \textbf{5.34} & \textbf{195.84} \\
& \hlcell{FORA ($N=2$) + TCC}
& \hlcell{0.5208}
& \hlcell{5.0678}
& \hlcell{1.57$\times$}
& \hlcell{\textbf{5.70}}
& \hlcell{5.57}
& \hlcell{194.57} \\
\midrule
\multirow{5}{*}{$512\times512$}
& DiT
& 4.6796 & 42.0977 & 1.00$\times$ & 4.98 & 6.56 & 186.18 \\
\cmidrule(lr){2-8}
& L2C
& 3.7654 & 32.7711 & 1.24$\times$ & 6.18 & 8.15 & 178.29 \\
& \hlcell{L2C + TCC}
& \hlcell{3.9908}
& \hlcell{33.9235}
& \hlcell{1.17$\times$}
& \hlcell{\textbf{6.03}}
& \hlcell{\textbf{6.94}}
& \hlcell{\textbf{178.56}} \\
\cmidrule(lr){2-8}
& FORA ($N=2$)
& 2.5709 & 21.0647 & 1.82$\times$ & 12.13 & 12.02 & 136.98 \\
& \hlcell{FORA ($N=2$) + TCC}
& \hlcell{2.6790}
& \hlcell{21.9778}
& \hlcell{1.75$\times$}
& \hlcell{\textbf{10.99}}
& \hlcell{\textbf{11.06}}
& \hlcell{\textbf{139.27}} \\
\bottomrule
\end{tabular}
\end{table}
Table~\ref{tab:imagenet_20_merged} summarizes the class-conditional results on
DiT-XL/2. On ImageNet $256\times256$, TCC improves both L2C and FORA, reducing
FID from 3.50 to 3.45 for L2C and from 6.82 to 5.70 for FORA. Compared with EOC,
TCC achieves the same FID and better sFID on L2C, and a lower FID on FORA. On
ImageNet $512\times512$, TCC consistently improves both cache baselines: it
reduces FID from 6.18 to 6.03 for L2C and from 12.13 to 10.99 for FORA. We omit
EOC at $512\times512$ because the original paper does not report this setting.
The improvements are most consistent on distributional image-quality metrics such as FID and sFID, while Inception Score remains broadly comparable across calibrated and uncalibrated cache variants.
Together with the PixArt-$\alpha$ results, these results show that TCC is effective across both text-to-image and
class-conditional generation, and across different cache reuse policies.

\subsection{Ablation Study}

\label{sec:ablation_online_tcc}
\begin{wrapfigure}{r}{0.45\textwidth}
\centering
\includegraphics[width=\linewidth]{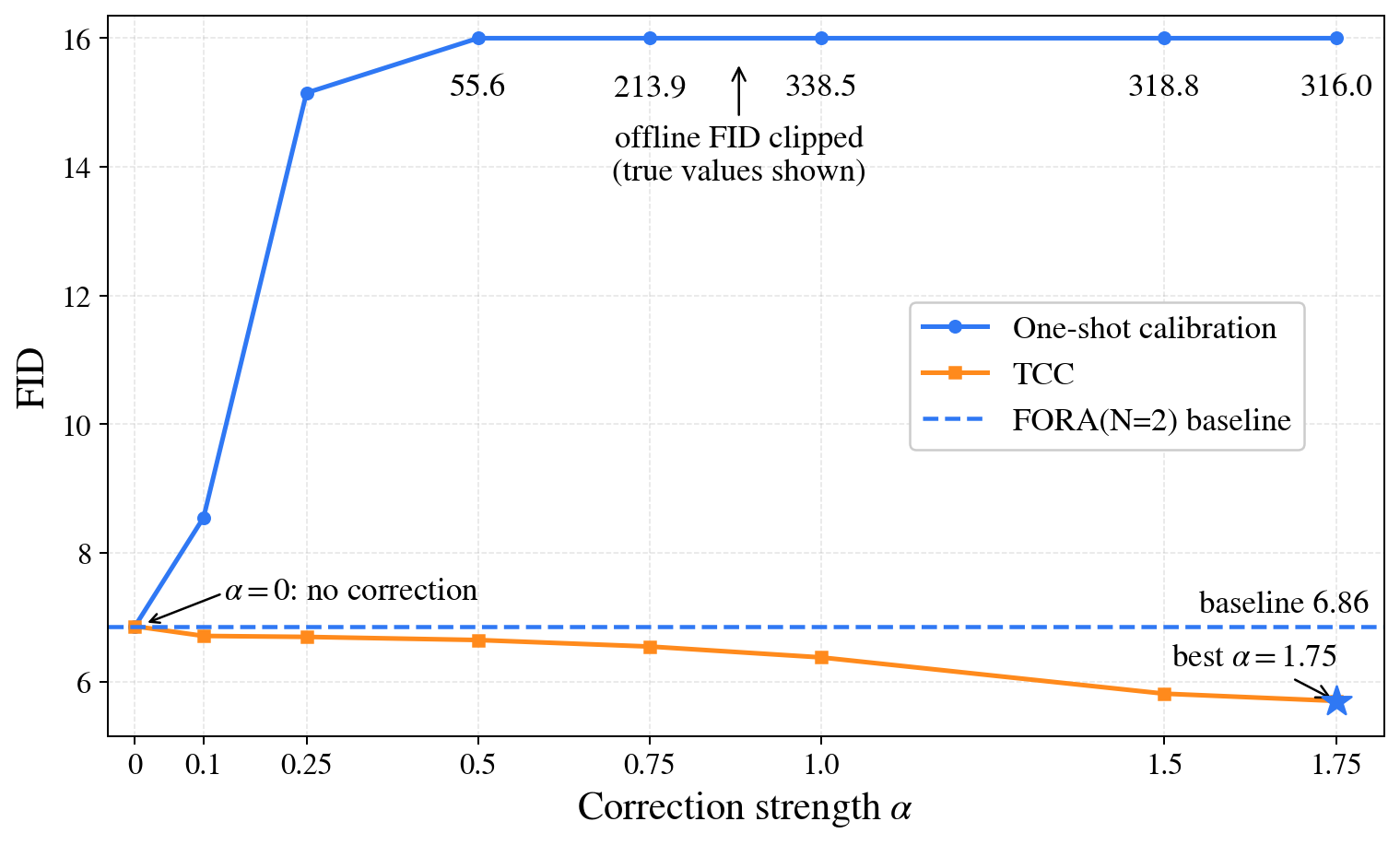}
\caption{
Sensitivity to the correction strength $\alpha$ on ImageNet $256\times256$ with
20-step DDIM sampling under FORA($N=2$). When $\alpha=0$, both variants reduce to
the FORA($N=2$) baseline. One-shot calibration becomes increasingly harmful as
$\alpha$ grows, while TCC remains stable and improves FID within the tested
range. For readability, one-shot points with very large FID are clipped and
marked.
}
\label{fig:alpha_sensitivity_online_offline}
\end{wrapfigure}
\textbf{Trajectory-consistent prior estimation.}
We first study whether calibration statistics should be estimated once from an
uncorrected cache trajectory or iteratively collected along the corrected
trajectory. This ablation is conducted on DiT-XL/2 at ImageNet $256\times256$
with 20-step DDIM sampling, using FORA($N=2$) as the cache baseline. We compare
a one-shot calibration variant, which estimates all calibration transforms from
the original uncorrected cache trajectory and keeps them fixed during sampling,
with TCC, which estimates each subsequent calibration transform after preceding
calibrations have been applied. Thus, TCC collects calibration statistics under
the trajectory distribution induced by earlier calibrations.
Figure~\ref{fig:alpha_sensitivity_online_offline} shows the sensitivity to the
correction strength $\alpha$. The two variants start from the same baseline at
$\alpha=0$. As $\alpha$ increases, one-shot calibration rapidly over-corrects the
cached representations, leading to severe degradation. This indicates that
calibration transformations estimated only once from the uncorrected trajectory do
not remain well matched after the sampling trajectory has been altered by prior
calibrations. In contrast, TCC remains stable and improves FID within the tested
range. This supports the main design of TCC: calibration statistics should be
estimated in a trajectory-consistent manner, rather than being fixed from a
single uncorrected cache trajectory. The corresponding numerical results are
provided in Appendix~\ref{app:online_vs_offline_tcc}.

\textbf{Effect of transformation components.}
\begin{table}[t]
\centering
\small
\setlength{\tabcolsep}{6pt}
\caption{
Ablation on transformation components under FORA($N=2$) on DiT-XL/2
(ImageNet $256\times256$, 20-step DDIM). For each calibration variant, we report the best
result over the tested $\alpha$ values, with the selected $\alpha$ shown in parentheses.
}
\label{tab:ablation_transform_components}
\begin{tabular}{lcc}
\toprule
\textbf{Method} & \textbf{FID $\downarrow$} & \textbf{sFID $\downarrow$} \\
\midrule
FORA($N=2$) baseline & 6.82 & 8.65 \\
Shift-only calibration ($\alpha=0.50$) & 6.81 & 8.87 \\
Scale+shift calibration ($\alpha=2.00$) & 5.81 & 6.63 \\
TCC ($\alpha=1.75$) & \textbf{5.70} & \textbf{5.57} \\
\bottomrule
\end{tabular}
\end{table}
We further ablate the transformation components used by TCC. Under the same
FORA($N=2$) setting, we compare three variants: shift-only calibration, which
only aligns the representation mean; scale+shift calibration, which aligns the mean and
a global scalar scale while removing the rotation component; and full TCC, which
uses the complete similarity transformation. As shown in
Table~\ref{tab:ablation_transform_components}, shift-only calibration barely
improves FID and slightly worsens sFID, indicating that mean alignment alone is
insufficient. Scale+shift calibration recovers a large portion of the FID gain,
but remains clearly worse than TCC in sFID. TCC with the complete similarity 
transformation achieves the best FID and
sFID trade-off, suggesting that the complete transformation is important for
preserving both image quality and structural fidelity. Full curves for FID and
sFID are provided in Appendix~\ref{app:transform_component_curves}.

\section{Related Work}
\textbf{Fast sampling for diffusion models.}
Diffusion inference can be accelerated by reducing the number of sampling steps or denoiser evaluations. Timestep-scheduling methods optimize sampling schedules along the denoising trajectory~\citep{OptimizedTimeSteps,AlignYourSteps}. Solver-based methods instead formulate generation as a reverse-time SDE or probability-flow ODE~\citep{ScoreSde}, leading to fast numerical samplers with fewer denoiser evaluations~\citep{pndm,dpmsolver,unipc,2024AMED-Solver}.
Training-based approaches learn few-step generative dynamics through distillation or consistency modeling~\citep{ProgressiveDistillation,ConsistencyModels}.

\textbf{Cache-based acceleration for diffusion models.}
Cache-based methods can be organized by the signals and granularity used for reuse decisions, including heuristic schedules, adaptive criteria, learned predictors, and fine-grained reuse policies.

Heuristic and fixed-schedule methods exploit temporal redundancy by reusing module outputs, block-level representations, or high-level features across denoising steps~\citep{fora,deepcache,blockcache}. Adaptive methods use proxy signals such as feature similarity, timestep embeddings, sample-dependent complexity, or sensitivity to estimate cacheability~\citep{SmoothCache,TeaCache,AdaCache,SenCache}.

Other methods learn or predict reuse decisions across layers, tokens, or timesteps~\citep{l2c,TokenCache}, refine token- or region-level reuse~\citep{toca,RegionAdaptive}, or optimize cache schedules across sampling steps~\citep{OmniCache,ECAD}. Overall, these works mainly improve reuse decisions, including when, where, and at what granularity representations are reused.

\textbf{Cache-induced error correction.}
After reuse decisions have been made, reused representations can still deviate from their full-computation counterparts and propagate errors through subsequent denoising steps, motivating correction before they are consumed by subsequent computation. Error-optimization and low-rank calibration methods estimate correction signals or structured transformations to compensate for discrepancies between reused and full-computation representations~\citep{eoc,2025icc}, but their priors are not explicitly estimated under the corrected history induced by previously applied corrections.

Gradient-optimized caching models progressive cache-induced perturbations and compensates for them through gradient propagation during inference~\citep{2025goc}, while cumulative-error minimization optimizes cache strategies with an offline cumulative-error model and dynamic programming under acceleration budgets~\citep{2026CEM}. These studies indicate that correcting reuse-induced discrepancies is distinct from designing cache schedules, while leaving room for calibration procedures that account for the corrected trajectory induced by previous corrections.

\section{Conclusion}
We have presented Trajectory-Consistent Calibration (TCC), a training-free calibration method for cache-accelerated diffusion inference.
TCC estimates local calibration priors from paired full-computation and cache-side representations along the corrected cache trajectory, matching each operator to its
inference-time history.
Experiments on PixArt-$\alpha$ and DiT-XL/2 show that TCC improves FID across representative cache strategies without modifying the denoiser, sampler, or base cache strategy.
These results suggest that trajectory-consistent calibration is a useful complement to cache-based
diffusion acceleration.
We discuss limitations and broader-impact considerations in Appendix~\ref{app:limitations}, with additional overhead analysis in Appendix~\ref{app:latency_overhead}.

% Anonymous submission: do not include acknowledgments.
% \begin{ack}
% \end{ack}

\clearpage
\bibliographystyle{plainnat}
\bibliography{references}

\clearpage
\appendix

\section{Technical Diagnostics}
\label{app:technical_diagnostics}

\subsection{Within-label dispersion during late denoising}
\label{app:within_label_dispersion}

Class-wise calibration relies on statistics estimated for each ImageNet label.
This implicitly assumes that, at a given denoising step and layer, the class-level
mean activation remains a reasonable representative of individual samples from
the same class. To examine when this assumption becomes weak, we measure the
within-label dispersion of DiT activations during sampling.

We follow the reverse DDIM order, where denoising proceeds from step 19 to step
0. Thus, smaller step indices correspond to later denoising stages. For a fixed
ImageNet class, we generate 100 samples and collect conditional activations from
each DiT block at every denoising step. For each branch and layer, we compute
the RMS of the within-label standard deviation across samples and compare it
with the RMS of the corresponding class-mean activation.

\begin{figure}[ht]
  \centering
  \includegraphics[width=\linewidth]{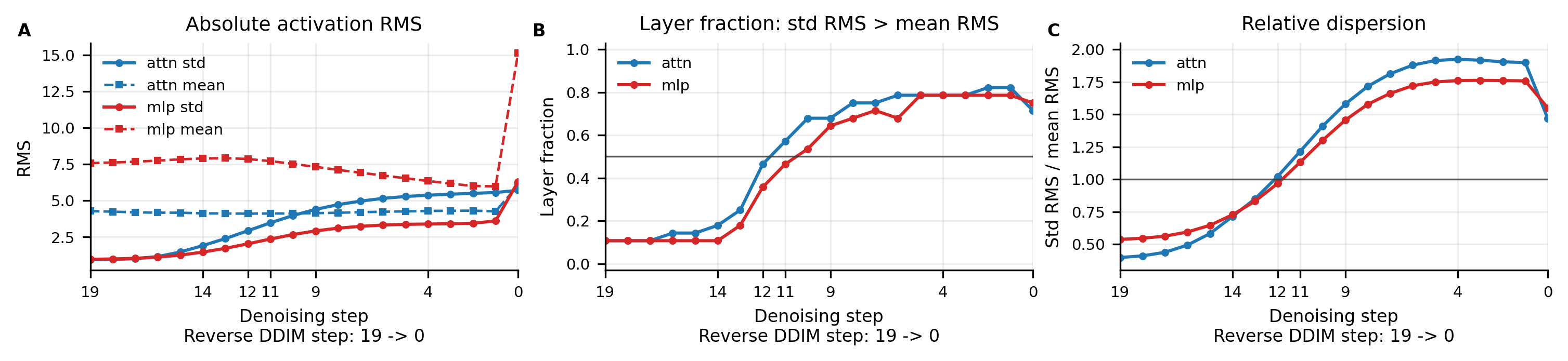}
  \caption{
\textbf{Within-label activation dispersion increases along the reverse denoising trajectory.}
We analyze 100 samples from the same ImageNet class and collect conditional
activations from DiT blocks at each DDIM step. The reverse trajectory proceeds
from step 19 to step 0.
\textbf{A:} Absolute RMS statistics averaged over layers; solid lines denote
within-label standard-deviation RMS, and dashed lines denote class-mean RMS.
\textbf{B:} Fraction of layers whose within-label standard-deviation RMS exceeds
their class-mean RMS. The horizontal line marks $50\%$ of layers.
\textbf{C:} Relative dispersion, measured as within-label standard-deviation RMS
divided by class-mean RMS. Values above 1 indicate that within-label variation
exceeds the class-mean activation scale.
}
  \label{fig:within_label_dispersion}
\end{figure}

As shown in Figure~\ref{fig:within_label_dispersion}, within-label dispersion
generally increases as denoising proceeds. Panel A reports the absolute RMS
statistics averaged over layers. In the early-to-middle stages, the class-mean
activation remains comparable to, or larger than, the within-label standard
deviation. However, the within-label standard deviation grows rapidly in later
steps. Panel B summarizes this trend by measuring the fraction of layers whose
within-label standard-deviation RMS exceeds the RMS of their corresponding class
mean. After the trajectory passes step 12 toward step 0, more than half of the
layers satisfy this condition. Panel C further shows the relative dispersion,
defined as the ratio between within-label standard-deviation RMS and class-mean
RMS. Values above 1 indicate that within-label variation exceeds the activation
scale of the corresponding class mean. The ratio rises sharply after the middle
steps and remains above 1 for most late denoising stages.

This observation supports our choice not to apply class-wise TCC uniformly
throughout the entire denoising trajectory. Under the DiT-256 20-step DDIM
setting, we restrict TCC to an early-to-middle window, typically steps 19--12.
The exact window may vary with the sampler, model, and caching schedule, and we
select it jointly with validation performance. Nevertheless, this statistic
provides a useful diagnostic: class-wise calibration is more reliable before
late-stage activations become dominated by sample-specific variation.

\subsection{Calibration window and hyperparameter selection}
\label{app:hyperparameter_selection}

TCC introduces two main hyperparameters: the calibration strength $\alpha$ and
the application window over denoising steps. We select them according to the
scale and granularity of the errors induced by each base cache strategy, while
keeping the denoiser, sampler, and cache schedule unchanged.

The calibration strength $\alpha$ controls how strongly the fitted statistical
transform is applied. Since different cache policies reuse representations at
different granularities, they induce representation deviations of different magnitudes.
For module-level cache methods such as FORA on DiT, reusing entire attention or
MLP outputs can tolerate a relatively stronger calibration. In contrast, learned
layer-level reuse or token-level reuse, as in L2C and ToCa, tends to require
smaller $\alpha$ values to avoid over-correction. The optimal strength can also
vary across model families; for PixArt-$\alpha$, we use more conservative
strengths for both FORA and ToCa. Within the same base method, moderate
$\alpha$ values are relatively stable across nearby resolutions and
sampling-step settings.

The application window is chosen to cover the early-to-middle part of the reverse
denoising trajectory. As discussed in Appendix~\ref{app:within_label_dispersion},
late denoising steps contain stronger sample-specific variation, making class- or
prompt-level calibration statistics less reliable. Therefore, we avoid applying
TCC uniformly to all steps and restrict calibration to a fixed early-to-middle
window. For 20-step DDIM sampling on DiT-XL/2, we typically apply TCC to steps
19--12; for the 512$\times$512 FORA setting, we use a slightly shorter window
19--14. For 50-step sampling, we use the corresponding relative early-to-middle
range, namely steps 49--30. For PixArt-$\alpha$ with 20-step DPM-Solver++ sampling,
we apply TCC to steps 19--14.

Table~\ref{tab:tcc_hyperparams} summarizes the hyperparameters used in the main
experiments and additional ablations. For all settings, within the selected application window, TCC is applied to
the cached attention outputs and MLP outputs of all transformer blocks. For
PixArt-$\alpha$, the attention outputs include both self-attention and
cross-attention. These settings are selected from a small pilot sweep and then
fixed for the reported evaluation. All prior statistics and calibration matrices
are computed in FP32.

\begin{table}[t]
\centering
\small
\setlength{\tabcolsep}{6pt}
\caption{
Summary of TCC hyperparameters used in the reported experiments. The denoiser,
sampler, and base cache schedule are kept unchanged in all settings.
}
\label{tab:tcc_hyperparams}
\begin{tabular}{llcc}
\toprule
\textbf{Model / setting} & \textbf{Base cache} & $\boldsymbol{\alpha}$ & \textbf{Window} \\
\midrule
PixArt-$\alpha$ 256$\times$256, 20-step & FORA $(N=3)$ & 0.50 & 19--14 \\
PixArt-$\alpha$ 256$\times$256, 20-step & ToCa $(N=3,R=90\%)$ & 0.25 & 19--14 \\
DiT-XL/2 256$\times$256, 20-step & FORA $(N=2)$ & 1.75 & 19--12 \\
DiT-XL/2 256$\times$256, 20-step & L2C & 0.50 & 19--12 \\
DiT-XL/2 512$\times$512, 20-step & FORA $(N=2)$ & 1.25 & 19--14 \\
DiT-XL/2 512$\times$512, 20-step & L2C & 0.50 & 19--12 \\
DiT-XL/2 256$\times$256, 50-step & FORA $(N=2)$ & 1.75 & 49--30 \\
DiT-XL/2 256$\times$256, 50-step & L2C / ToCa & 0.50 & 49--30 \\
\bottomrule
\end{tabular}
\end{table}

\section{Additional Experimental Details and Ablations}
\label{app:additional_experiments}

\subsection{Numerical results for one-shot calibration and TCC}
\label{app:online_vs_offline_tcc}

Table~\ref{tab:online_vs_offline_tcc} reports representative points from the
coarse $\alpha$ sweep used in Figure~\ref{fig:alpha_sensitivity_online_offline}.
When $\alpha=0$, both variants reduce to the original FORA($N=2$) baseline,
since no correction is applied. Even with a weak correction strength, one-shot
calibration already degrades generation quality. As $\alpha$ increases, the
degradation becomes severe, indicating that a fixed calibration transform
estimated from the uncorrected trajectory is not compatible with the trajectory
shifted by previous calibrations. In contrast, TCC remains stable and improves
FID as the correction strength increases within the tested range.

\begin{table}[ht]
\centering
\caption{
Comparison between one-shot calibration and TCC on ImageNet $256\times256$ with
20-step DDIM sampling under FORA($N=2$). All variants apply calibration to all
cached attention and MLP outputs.
}
\label{tab:online_vs_offline_tcc}
\small
\setlength{\tabcolsep}{6pt}
\begin{tabular}{lccc}
\toprule
Method & $\alpha$ & FID $\downarrow$ & sFID $\downarrow$ \\
\midrule
FORA($N=2$) & 0 & 6.82 & 8.65 \\
\midrule
FORA($N=2$) + One-shot calibration & 0.1 & 8.55 & 14.69 \\
\rowcolor{gray!15}
FORA($N=2$) + TCC & 0.1 & 6.71 & 8.77 \\
\midrule
FORA($N=2$) + One-shot calibration & 0.5 & 55.61 & 198.80 \\
\rowcolor{gray!15}
FORA($N=2$) + TCC & 0.5 & 6.65 & 8.78 \\
\midrule
FORA($N=2$) + One-shot calibration & 1.5 & 318.75 & 225.75 \\
\rowcolor{gray!15}
FORA($N=2$) + TCC & 1.5 & 5.81 & 6.51 \\
\bottomrule
\end{tabular}
\end{table}

\subsection{Full curves for transformation-component ablation}
\label{app:transform_component_curves}

This subsection provides the full results for the transformation-component
ablation discussed in the main experimental section. In the main text, we report
a compact comparison among shift-only calibration, scale+shift calibration, and
the complete TCC transformation. Here, we further show the complete FID and sFID
curves under different correction strengths $\alpha$.

\begin{figure}[ht]
    \centering
    \begin{subfigure}[t]{0.48\linewidth}
        \centering
        \includegraphics[width=\linewidth]{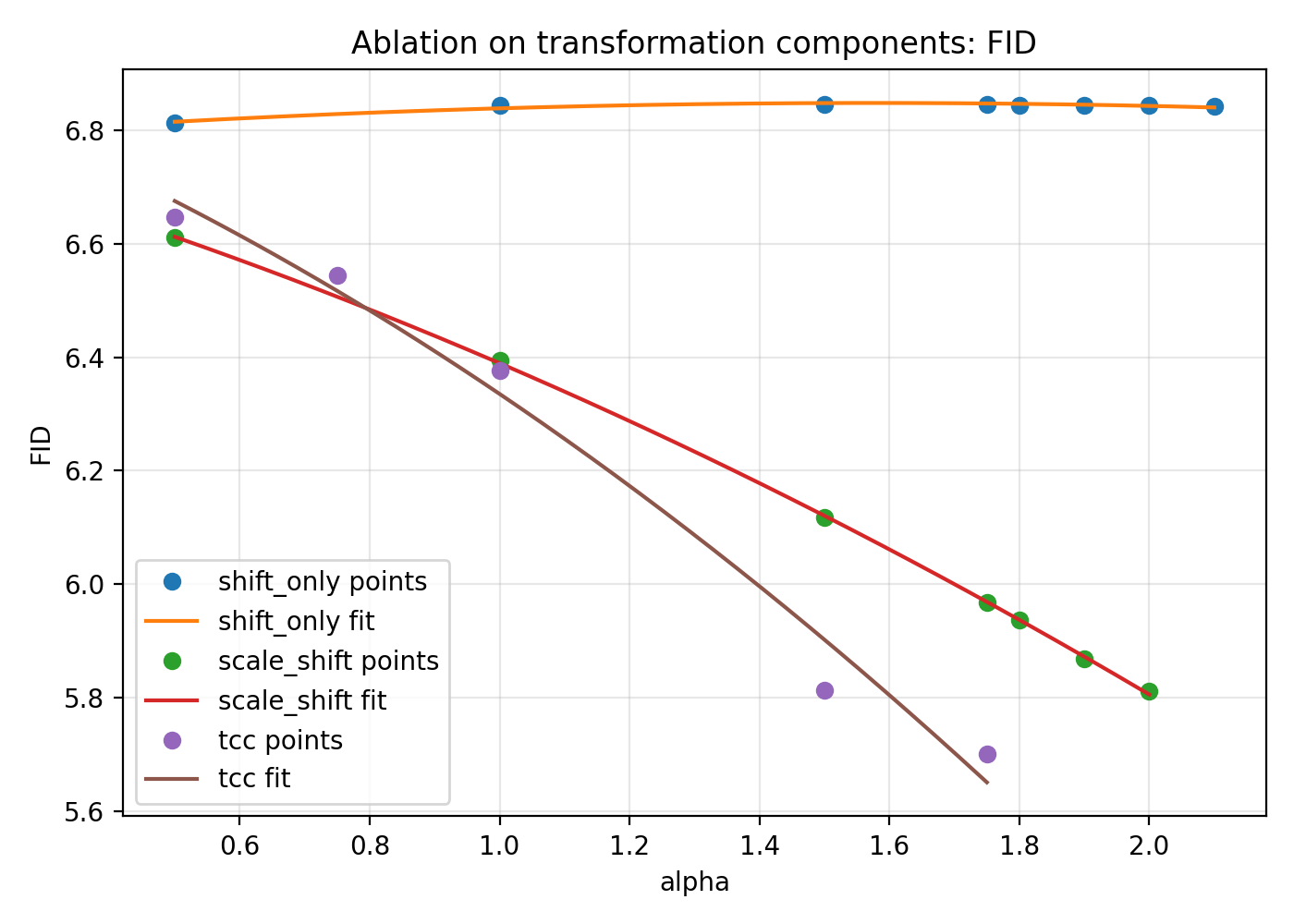}
        \caption{FID versus $\alpha$.}
        \label{fig:ablation_transform_fid}
    \end{subfigure}
    \hfill
    \begin{subfigure}[t]{0.48\linewidth}
        \centering
        \includegraphics[width=\linewidth]{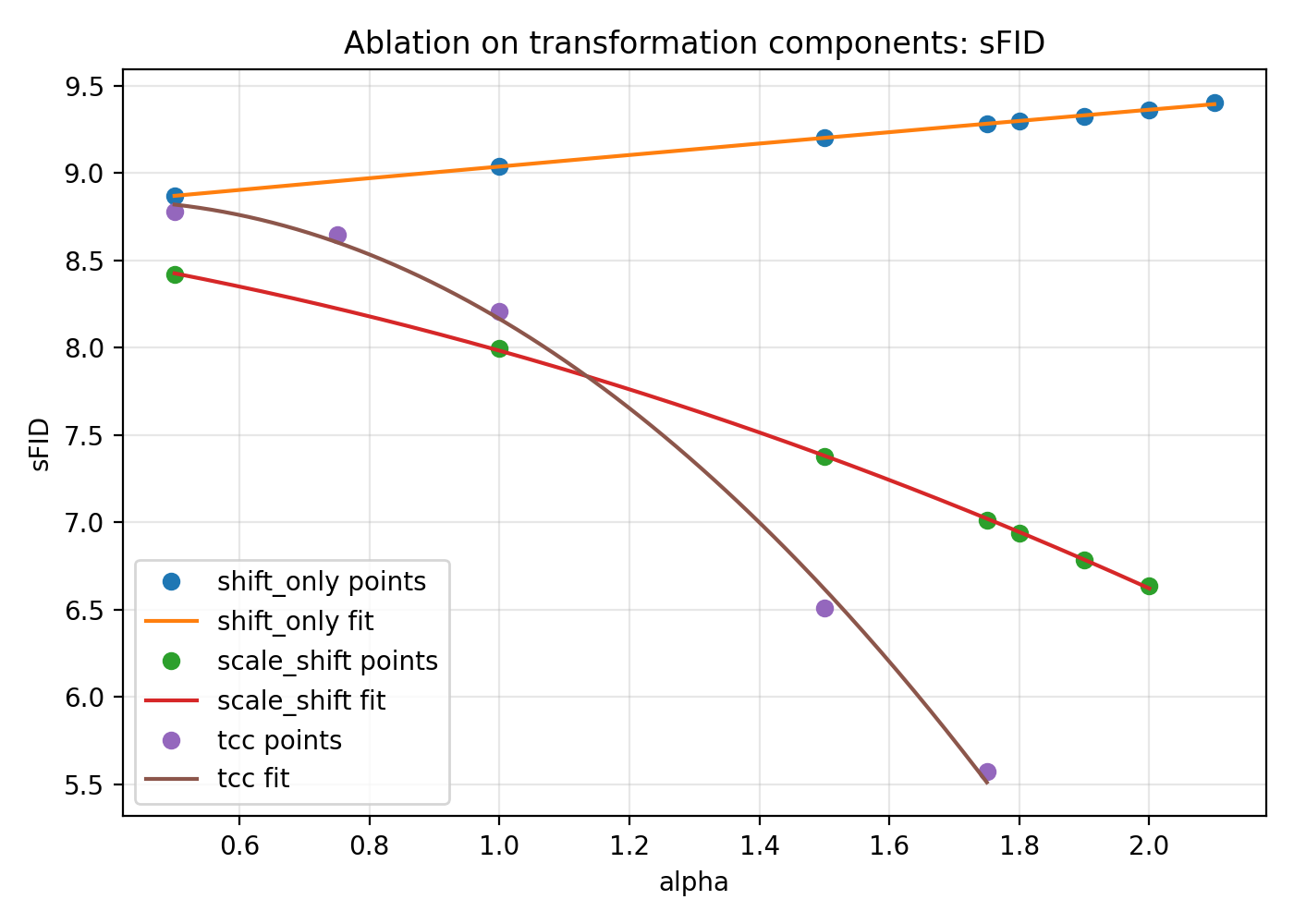}
        \caption{sFID versus $\alpha$.}
        \label{fig:ablation_transform_sfid}
    \end{subfigure}

    \caption{
    Transformation-component ablation under FORA($N=2$). Markers denote measured
    results, and curves are quadratic fits for visualizing the trend. Shift-only
    calibration remains close to the cache baseline, while scale+shift calibration
    substantially improves FID and approaches TCC. Compared with scale+shift,
    full TCC achieves a much lower sFID, showing that the complete transformation
    is more effective in preserving structural fidelity.
    }
    \label{fig:ablation_transform_curves}
\end{figure}

Since different calibration forms may reach their best FID and sFID at different
correction strengths, a single $\alpha$ value may not fully reflect their
respective behavior. Therefore, we evaluate a range of commonly used $\alpha$
values under the same experimental setting and plot the resulting curves in
Figures~\ref{fig:ablation_transform_fid} and~\ref{fig:ablation_transform_sfid}.
All variants are evaluated on DiT-XL/2 at ImageNet $256\times256$ with
20-step DDIM sampling, using FORA($N=2$) as the cache baseline. TCC is applied
to the cached attention and MLP outputs in the same application window and layer
range as in the main ablation.

The results are consistent with the compact table in the main text. Shift-only
calibration is clearly insufficient, indicating that mean alignment alone cannot
adequately correct the deviation introduced by representation reuse. Scale+shift
calibration recovers a large portion of the FID improvement and reaches a FID
close to the complete TCC transformation. However, its sFID remains noticeably
worse than TCC. In contrast, the complete TCC transformation achieves the best
overall trade-off, with both the lowest FID and a substantially better sFID. This
suggests that the rotation component is important for preserving structural
statistics, and supports the necessity of using the full TCC calibration rather
than only shift or scale+shift alignment.

\subsection{Pooling strategy for calibration statistics}
\label{app:pooling_strategy}
We ablate how calibration statistics are pooled when constructing the TCC
calibration transforms. Table~\ref{tab:ablation_pool} compares three pooling
strategies on ImageNet $256\times256$ under different correction strengths.
Across all settings, class-pool performs much worse than token-aware variants,
suggesting that collapsing calibration statistics into class-level averages
discards token-specific spatial and structural variations and can lead to severe
over-correction.

Introducing token-level statistics substantially improves performance. At
$\alpha=1.75$, mixed token-wise pooling reduces FID from 20.43 to 7.34, while
token-pool further improves it to 5.70. The same trend also holds at
$\alpha=1.50$, where token-pool achieves the best FID, sFID, and IS among all
pooling strategies. These results indicate that preserving fine-grained
token-level statistics provides a more reliable calibration direction for cached
representations.

\begin{table}[t]
\centering
\caption{
Ablation on pooling strategy for TCC under different correction strengths. All
results are evaluated on ImageNet $256\times256$ and use all transformer layers.
}
\label{tab:ablation_pool}
\begin{tabular}{lcccc}
\toprule
$\alpha$ & Pooling strategy & FID $\downarrow$ & sFID $\downarrow$ & IS $\uparrow$ \\
\midrule
1.25 & Class-pool & 15.05 & 33.93 & 132.78 \\
1.25 & Mixed &8.58 &13.81 &176.46 \\
1.25 & Token-pool & \textbf{6.13} & \textbf{7.20} & \textbf{194.32} \\
\midrule
1.50 & Class-pool & 17.37 & 40.48 & 121.20  \\
1.50 & Mixed & 8.00 & 10.94 & 179.21 \\
1.50 & Token-pool & \textbf{5.81} & \textbf{6.27} & \textbf{196.72} \\
\midrule
1.75 & Class-pool & 20.43 & 49.43 & 107.20  \\
1.75 & Mixed & 7.34 & 8.08 & 182.11  \\
1.75 & Token-pool & \textbf{5.70} & \textbf{5.57} & \textbf{194.57} \\
\bottomrule
\end{tabular}
\end{table}

\subsection{Additional ImageNet results with 50-step DDIM sampling}
\label{app:imagenet_50step}

We further evaluate TCC on ImageNet class-conditional generation with
DiT-XL/2 under the $256\times256$, 50-step DDIM setting. This setting allows us
to compare with L2C, FORA, and ToCa under the same benchmark. In contrast, ToCa
does not report a tuned token-caching ratio for the 20-step setting, and directly
reusing the reported ratio leads to a poor baseline. Therefore, we use the
20-step setting in the main paper for L2C and FORA, and provide the 50-step
comparison here as an additional result.

\begin{table}[ht]
\centering
\caption{
Class-conditional image generation on ImageNet $256\times256$ with 50-step DDIM
sampling. For ToCa, we use $R=93\%$; TCC hyperparameters are summarized in
Table~\ref{tab:tcc_hyperparams}.
}
\label{tab:imagenet256_50}
\small
\setlength{\tabcolsep}{4pt}
\begin{tabular}{lcccccccc}
\toprule
Method & Latency (s) $\downarrow$ & FLOPs $\downarrow$ & Speed $\uparrow$ & FID $\downarrow$ & sFID $\downarrow$ & IS $\uparrow$ & Precision $\uparrow$ & Recall $\uparrow$ \\
\midrule
DiT & 2.0435 & 23.7637 & 1.00$\times$ & 2.26 & 4.32 & 238.21 & 0.81 & 0.60 \\
\midrule
L2C & 1.6494 & 18.1264 & 1.24$\times$ & 2.25 & \textbf{4.28} & \textbf{246.47} & 0.81 & 0.58 \\
\rowcolor{gray!15}
L2C + TCC & 1.7638 & 18.8494 & 1.16$\times$ & \textbf{2.22} & 4.32 & 245.16 & 0.81 & 0.59 \\
\midrule
FORA($N=2$) & 1.1929 & 11.9085 & 1.71$\times$ & 2.65 & 4.71 & \textbf{238.04} & 0.80 & 0.59 \\
\rowcolor{gray!15}
FORA($N=2$) + TCC & 1.3163 & 12.6695 & 1.55$\times$ & \textbf{2.49} & \textbf{4.69} & 237.00 & 0.80 & 0.59 \\
\midrule
ToCa($N=2$) & 1.5386 & 13.5404 & 1.33$\times$ & 2.60 & \textbf{4.43} & \textbf{236.57} & 0.80 & 0.58 \\
\rowcolor{gray!15}
ToCa($N=2$) + TCC & 1.6186 & 14.3775 & 1.26$\times$ & \textbf{2.55} & 4.49 & 235.84 & 0.80 & 0.59 \\
\bottomrule
\end{tabular}
\end{table}

As shown in Table~\ref{tab:imagenet256_50}, TCC improves FID for all three cache
baselines in this setting. For L2C, TCC reduces FID from 2.25 to 2.22, while
keeping precision unchanged and slightly improving recall from 0.58 to 0.59. For
FORA($N=2$), TCC reduces FID from 2.65 to 2.49 and slightly improves sFID from
4.71 to 4.69. For ToCa($N=2$), TCC reduces FID from 2.60 to 2.55, with a slight
increase in sFID. These results provide additional evidence that TCC can be applied to different
cache schedules and acceleration mechanisms beyond the 20-step main setting.

\subsection{Additional qualitative comparisons}
\label{app:qualitative}

We provide additional qualitative comparisons to further illustrate the visual
effect of TCC under cache-accelerated sampling. These examples complement the
quantitative results in the main paper by showing how TCC affects generated
samples under matched prompts or class labels, random seeds, and sampling
configurations.

\begin{figure}[H]
    \centering
    \includegraphics[width=\linewidth]{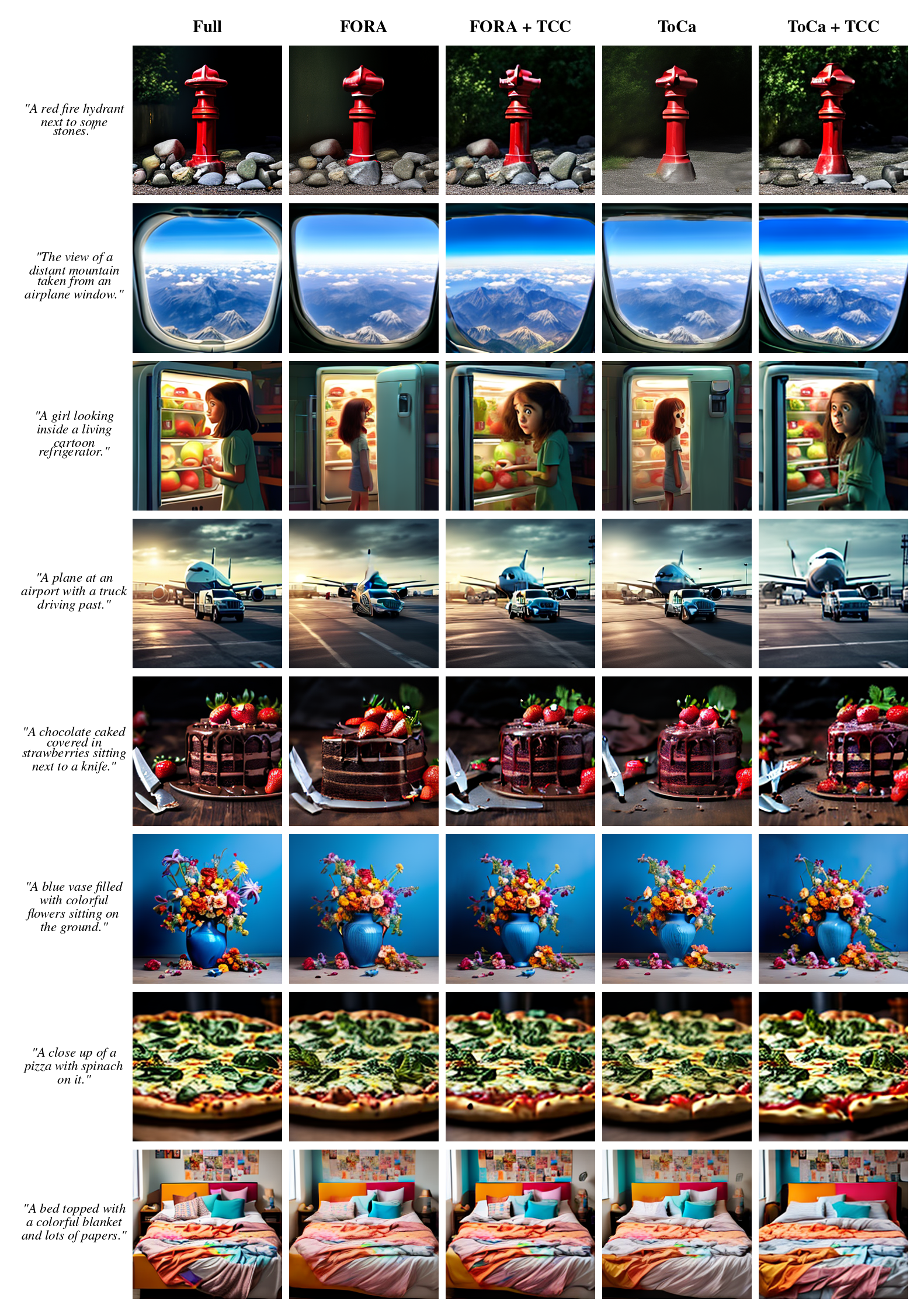}
    \caption{
    Additional qualitative comparisons on PixArt-$\alpha$ under cache-accelerated
    text-to-image generation. All comparisons use matched prompts, random seeds,
    and sampling settings. TCC preserves the overall composition while improving
    prompt-relevant details and mitigating local degradation caused by cache reuse.
    }
    \label{fig:app_pixart_more}
\end{figure}

For text-to-image generation, we show more PixArt-$\alpha$ comparisons under the
same setting as Fig.~\ref{fig:pixart_qualitative}. Specifically, we compare the
base cache-accelerated sampler with its TCC-calibrated counterpart using matched
MS-COCO prompts, seeds, and 20-step DPM-Solver++ sampling at $256\times256$
resolution. As shown in Fig.~\ref{fig:app_pixart_more}, TCC generally preserves
the overall layout of the cached samples while improving prompt-relevant details
and reducing local visual degradation introduced by cache reuse.

\begin{figure}[H]
    \centering
    \includegraphics[width=0.95\linewidth]{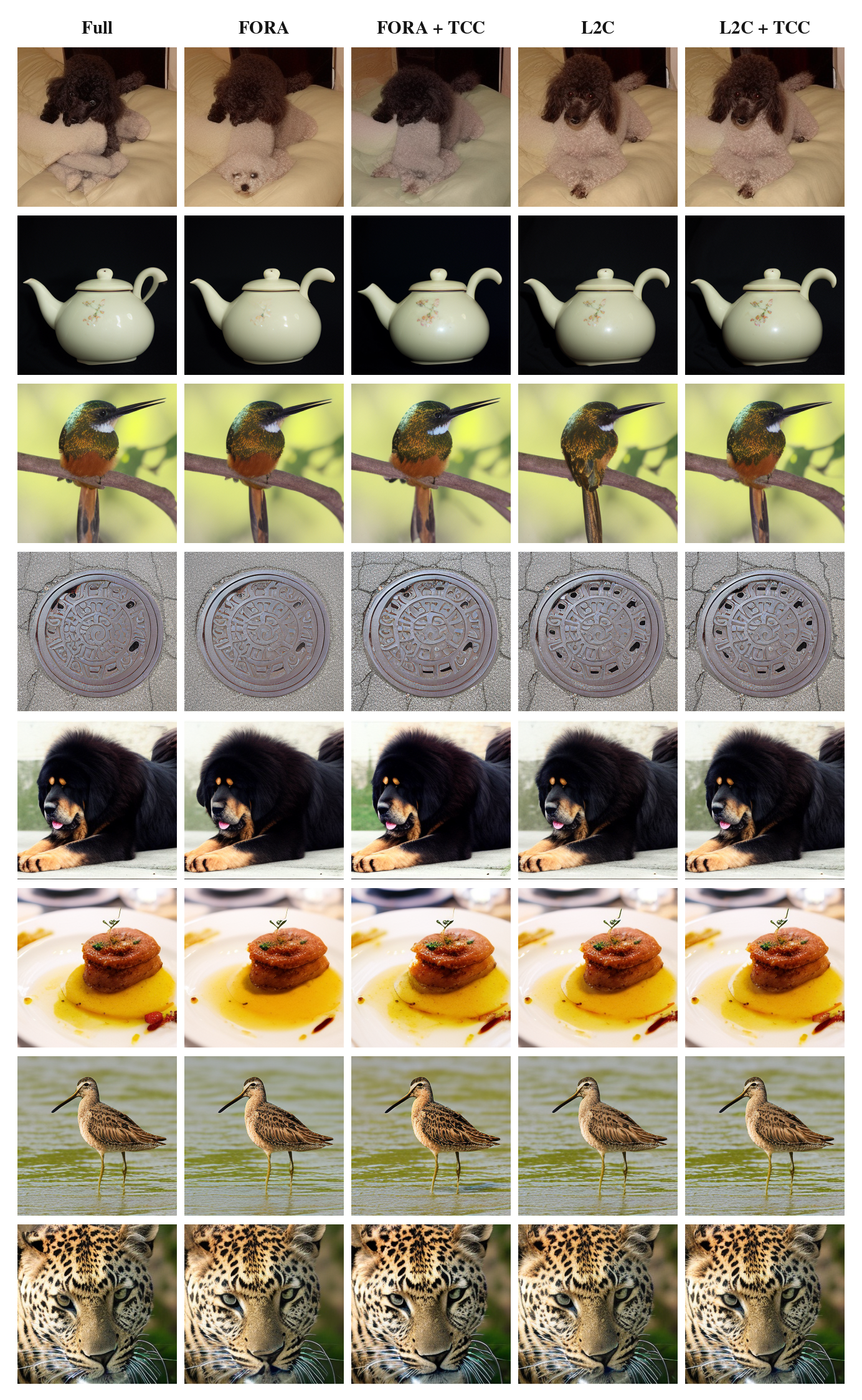}
    \caption{
    Additional qualitative comparisons on DiT-XL/2 for ImageNet class-conditional
    generation at $256\times256$ resolution with 20-step DDIM sampling and
    classifier-free guidance scale 4.0. All comparisons use matched class labels,
    random seeds, and sampling settings.
    }
    \label{fig:app_dit_256_cfg4}
\end{figure}

For class-conditional generation, we additionally visualize DiT-XL/2 samples on
ImageNet at $256\times256$ resolution with 20-step DDIM sampling and
classifier-free guidance scale 4.0. We compare the corresponding cache baseline
and TCC-calibrated results under matched class labels and random seeds. As shown
in Fig.~\ref{fig:app_dit_256_cfg4}, TCC improves visual fidelity in several
cases, especially by reducing artifacts and recovering finer object or texture
details, while maintaining the semantic category of the generated image.
These qualitative results are consistent with the quantitative improvements
reported in the main paper and suggest that TCC improves cache-accelerated
generation not only in distribution-level metrics such as FID, but also in
sample-level visual quality.

\subsection{Number of prompts for PixArt representative selection}
\label{app:pixart_prompt_number}

We further study how many text prompts are needed to estimate representative
TCC statistics for text-to-image generation. In this experiment, PixArt-$\alpha$
is used as the base model, ToCa is used as the cache acceleration method, and
TCC is applied as a calibration module. We fix the ToCa configuration to $N=3$
and $R=90\%$, and only vary the number of MS-COCO2017 captions used for representative
selection. To isolate the effect of prompt count, all settings use the same
correction strength $\alpha=0.25$ and the same application window 19--14.

Table~\ref{tab:ablation_pixart_prompt_num} compares the results obtained with
1k, 2k, and 5k representative prompts. The three settings yield similar FID and
CLIP scores, indicating that the estimated TCC statistics are not sensitive to
the prompt-set size within this range.

\begin{table}[ht]
\centering
\small
\setlength{\tabcolsep}{6pt}
\caption{
Ablation on the number of representative prompts used to estimate TCC statistics
for PixArt-$\alpha$ under the same ToCa-accelerated sampling pipeline.
}
\label{tab:ablation_pixart_prompt_num}
\begin{tabular}{lcc}
\toprule
\multirow{2}{*}{\textbf{Prompts}}
& \multicolumn{2}{c}{\textbf{MS-COCO2017}} \\
\cmidrule(lr){2-3}
& \textbf{FID-30K} $\downarrow$ & \textbf{CLIP} $\uparrow$ \\
\midrule
1k & \textbf{27.36} & \textbf{30.37} \\
2k & 28.13 & 30.25 \\
5k & 27.79 & 30.33 \\
\bottomrule
\end{tabular}
\end{table}

The results show that increasing the representative prompt set from 1k to 5k
does not lead to consistent improvements. The 1k setting performs slightly best
among the tested choices, suggesting that a relatively small prompt set is
already sufficient for estimating stable TCC statistics in this setting. We
therefore use 1k representative prompts for PixArt experiments unless otherwise
specified.

\subsection{\texorpdfstring{Latency overhead of ToCa on PixArt-$\alpha$}{Latency overhead of ToCa on PixArt-alpha}}
\label{app:latency_overhead}

\paragraph{Latency and FLOPs measurement.}
Latency is measured with batch size 8 using CUDA events. The latency values
reported in the tables correspond to the end-to-end wall-clock time for generating
one batch of eight images. For each method, we first perform one warm-up sampling
run and then report the average latency over five timed sampling runs under the
same sampling configuration. The speedup values are computed from these batch
latency measurements relative to the corresponding full-computation baseline.

The FLOPs reported in the tables are normalized per generated image. They are
computed separately from the CUDA-event timing: for DiT and PixArt-$\alpha$
cache/TCC variants, FLOPs are obtained from analytical estimators matched to the
corresponding cache schedule and TCC application window; for the DiT-ToCa
baseline, FLOPs are taken from the original ToCa FLOPs counter. Computing FLOPs
separately avoids adding profiling overhead to the latency measurements and makes
the FLOPs numbers reflect estimated dense model-computation cost rather than
measured runtime. Since all methods within each table are compared under the same
batch-size and FLOPs-normalization protocol, these numbers are intended to support
relative speed--cost comparisons rather than to equate wall-clock latency and
FLOPs units directly.

The FLOPs estimates include the denoiser forward computations and the additional
TCC calibration transforms, while excluding VAE decoding, text encoding, I/O, and
metric computation. The additional FLOPs introduced by TCC come only from applying
the fitted calibration transforms at selected cached sites. For cache baselines,
reused cached representations are counted as zero additional dense-computation
FLOPs, and only the freshly computed modules or tokens are counted according to
the corresponding cache schedule.

\paragraph{Latency analysis of ToCa+TCC.}
The end-to-end latency numbers in Table~\ref{tab:pixart_coco} are measured under
the above protocol and are used for the main speed-quality comparison. To further
explain the apparently unusual latency behavior of ToCa+TCC, especially the fact
that it can be slower in wall-clock time despite lower estimated FLOPs, we conduct
an additional fine-grained profiling experiment on a single sampling pass. This
extra profiling run is not used to report the main latency numbers; it is used
only to analyze the relative composition of ToCa's runtime overhead.

As shown in Table~\ref{tab:pixart_coco}, ToCa reduces the estimated model
FLOPs on PixArt-$\alpha$, but the corresponding wall-clock acceleration is
limited. This discrepancy mainly comes from two sources. First, ToCa performs
token-level reuse rather than whole-module reuse. Even under aggressive reuse,
fresh tokens are still recomputed in the cross-attention and MLP modules, so a
non-trivial portion of transformer computation remains. Second, ToCa introduces
a dynamic token-management path, including token scoring, freshness
reweighting, sorting, token gathering/scattering, and cache-index updates.
These operations contribute modestly to model FLOPs, but can introduce
noticeable wall-clock overhead due to indexing, memory movement, synchronization,
and many small GPU kernels.

For this analysis, we further instrument ToCa-r90 on PixArt-$\alpha$ at
$256\times256$ resolution with 20 sampling steps using CUDA events. This profiling is used only to analyze the relative latency
composition. Because module-level profiling requires additional CUDA events and
synchronization points, the profiled absolute latency is not intended to match
the end-to-end latency reported in Table~\ref{tab:pixart_coco}.

\begin{figure}[H]
\centering
\includegraphics[width=0.92\linewidth]{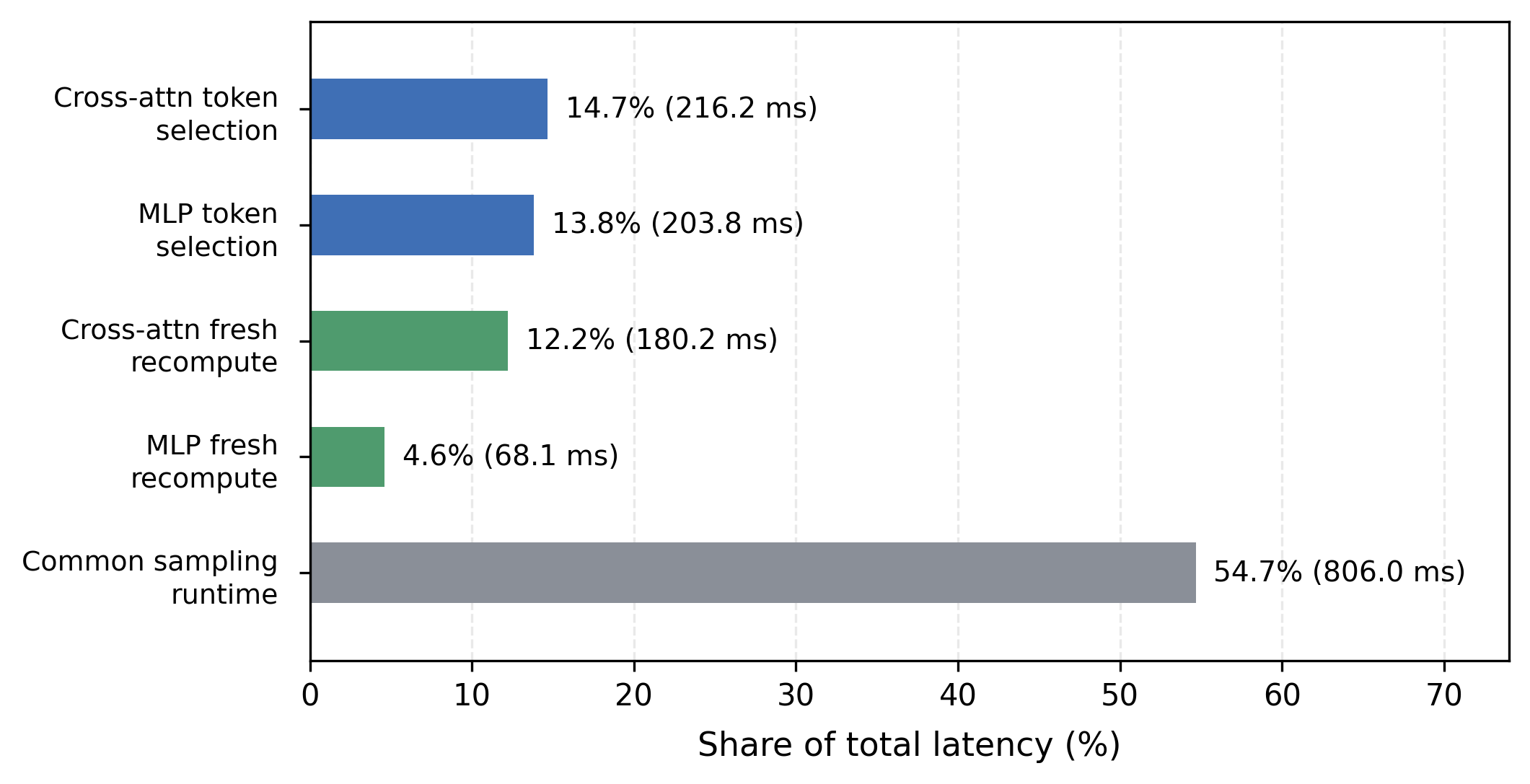}
\caption{
Latency breakdown of ToCa-r90 on PixArt-$\alpha$ at $256\times256$ resolution
with 20 sampling steps. Blue bars denote dynamic token-selection overhead,
green bars denote fresh-token recomputation in cached modules, and the gray bar
denotes common sampling runtime, including regular forward computation,
residual normalization, cache reuse operations, and other non-ToCa-specific
sampling costs. Percentages are normalized by the total profiled latency.
}
\label{fig:toca_latency_breakdown}
\end{figure}

Figure~\ref{fig:toca_latency_breakdown} shows the measured latency composition.
In this profiling run, ToCa-r90 takes $1474.4$ ms per batch in total. The dynamic
token-selection path accounts for $420.1$ ms, including token selection before
cross-attention cache computation and before MLP cache computation:
\[
\texttt{cross\_token\_select}=216.2\text{ ms},\quad
\texttt{mlp\_token\_select}=203.8\text{ ms}.
\]
Together, these two selection stages contribute $28.5\%$ of the total profiled
latency. Inside this dynamic path, the main measured components include score
evaluation ($146.7$ ms, $9.9\%$), spatial bonus computation ($71.2$ ms,
$4.8\%$), sorting ($41.6$ ms, $2.8\%$), and cache-index maintenance
($39.3$ ms, $2.7\%$). These costs are not well captured by model-FLOPs
counting, since they are dominated by memory access, indexing, synchronization,
and small-kernel overheads rather than dense matrix multiplication.

The dynamic selection path, however, is not the only reason for the limited
latency speedup. As shown in Figure~\ref{fig:toca_latency_breakdown},
fresh-token recomputation in cached modules also takes a substantial fraction of
runtime: $180.2$ ms ($12.2\%$) for cross-attention and $68.1$ ms ($4.6\%$) for
MLP. Therefore, even if the measured selection overhead were removed, the
ToCa-r90 latency would decrease from $1474.4$ ms to about $1054.3$ ms, which is
still not close to the acceleration suggested by FLOPs reduction alone. This is
because ToCa still performs partial token computation in attention and MLP
modules, whereas static cache methods such as FORA reuse whole module outputs
on cache steps.

Overall, the latency gap of ToCa on PixArt-$\alpha$ comes from the combination
of retained fresh-token computation and dynamic token-management overhead. This
explains why ToCa has wall-clock latency close to the native PixArt-$\alpha$
baseline in Table~\ref{tab:pixart_coco}, despite reducing the estimated FLOPs.
TCC further applies a calibration transform to cached representations. Although
this adds less than $10\%$ extra FLOPs over the ToCa-r90 setting, the base ToCa
latency is already close to the native baseline. Therefore, ToCa+TCC can become
slower than native PixArt-$\alpha$ in wall-clock time. The ToCa+TCC result on
PixArt-$\alpha$ should thus be interpreted as a quality-recovery setting under
aggressive token-level reuse, with an explicit latency trade-off.

\subsection{Compute resources}
\label{app:compute_resources}

The reported experiments were conducted on NVIDIA H800 GPUs with 80GB memory. We report approximate compute in H800 GPU-hours, computed as wall-clock time multiplied by the number of GPUs. For our DiT-XL/2 ImageNet evaluations, 50K-sample 20-step TCC runs take about 1--2 GPU-hours per method setting at $256\times256$ and about 6--8 GPU-hours at $512\times512$. The corresponding offline prior construction takes about 2 GPU-hours at $256\times256$ and up to about 25 GPU-hours for the largest $512\times512$ setting. For PixArt-$\alpha$, MS-COCO-30K sampling runs take about 1--2 GPU-hours per method setting, and prior-construction runs, including prompt-count ablations up to 5K prompts, take about 1 GPU-hour per method pack.

For storage overhead, in the DiT-XL/2 $256\times256$ 20-step FORA($N=2$) setting, the stored calibration operators for one calibrated cache timestep occupy about 0.3GB. This size is determined mainly by the DiT backbone and the selected calibration sites, rather than by the particular base cache policy. For example, four calibrated cache timesteps within the 19--12 apply window require about 1.2GB of stored calibration operators in the DiT-XL/2 $256\times256$ 20-step FORA($N=2$) setting. This calibration pack is constructed with 5 H800 GPUs in parallel.

\section{Limitations and Broader Impact}
\label{app:limitations}

TCC requires an offline prior-estimation stage and additional calibration transforms during cached inference, which introduce extra FLOPs and can reduce wall-clock speedup when the underlying cache method already incurs dynamic token-selection or memory-movement overhead; a detailed overhead analysis is provided in Appendix~\ref{app:latency_overhead}.
Calibration strength and sites are chosen empirically per base cache strategy, reflecting strategy-dependent error magnitudes and granularities.
Although TCC consistently improves FID, gains do not transfer uniformly to all metrics; CLIP Score or sFID can be slightly less favorable in some settings.
By improving quality preservation in cache-accelerated image generation, TCC may reduce compute and energy costs, but may also lower the cost of misusing generative models for deceptive or harmful synthetic content.
\newpage

\end{document}